\documentclass[conference]{IEEEtran}
\IEEEoverridecommandlockouts
\usepackage{booktabs}
\usepackage{cite}
\usepackage{amsmath,amssymb,amsfonts}
\usepackage{algorithmic}
\usepackage{graphicx}
\usepackage{textcomp}
\usepackage{xcolor}
\def\BibTeX{{\rm B\kern-.05em{\sc i\kern-.025em b}\kern-.08em
    T\kern-.1667em\lower.7ex\hbox{E}\kern-.125emX}}
\begin{document}

\title{DAPONet: A Dual Attention and Partially Overparameterized Network for Real-Time Road Damage Detection\\
}
\author{
\IEEEauthorblockN{Weichao Pan\textsuperscript{1} \quad Jiaju Kang\textsuperscript{1,2,*}\thanks{*Corresponding author: kjj\_python@163.com} \quad Xu Wang\textsuperscript{1}, \quad Zhihao Chen\textsuperscript{3} \quad Yiyuan Ge\textsuperscript{3} }

\IEEEauthorblockA{\textsuperscript{1} Shandong Jianzhu University \hspace{4mm} \textsuperscript{2} Beijing Normal University} \hspace{4mm} \textsuperscript{3} BISTU

}

\maketitle

\begin{abstract}
Current road damage detection methods, relying on manual inspections or sensor-mounted vehicles, are inefficient, limited in coverage, and often inaccurate, especially for minor damages, leading to delays and safety hazards. To address these issues and enhance real-time road damage detection using street view image data (SVRDD), we propose DAPONet, a model incorporating three key modules: a dual attention mechanism combining global and local attention, a multi-scale partial over-parameterization module, and an efficient downsampling module. DAPONet achieves a mAP50 of 70.1\% on the SVRDD dataset, outperforming YOLOv10n by 10.4\%, while reducing parameters to 1.6M and FLOPs to 1.7G, representing reductions of 41\% and 80\%, respectively. On the MS COCO2017 val dataset, DAPONet achieves an mAP50-95 of 33.4\%, 0.8\% higher than EfficientDet-D1, with a 74\% reduction in both parameters and FLOPs. 
\end{abstract}

\begin{IEEEkeywords}
SVRDD, Road Damage Detection, Real-Time Object Detection, Global and Local Attention, Partial Over-Parameterization
\end{IEEEkeywords}

\section{Introduction}
The maintenance and management of roads\cite{b1} are critical to the safety and operational efficiency of cities. However, current road damage detection\cite{b2}\cite{b3} methods mainly rely on manual inspections or vehicle-mounted sensors for data collection, and these methods have many limitations, such as low detection efficiency, high cost, and complex data processing. In addition, the lack of accuracy of traditional methods in detecting minor damages (e.g., small cracks or initial potholes) makes it difficult to detect and repair potential road problems in a timely manner, increasing the risk of traffic accidents.

Recent advancements in road damage detection have focused on improving model accuracy, robustness, and efficiency. Key developments include enhancements to deep learning architectures, integration of attention mechanisms, lightweight model design, and adaptability across various environments\cite{b4}-\cite{new6}.
Improved versions of deep learning models like YOLOX-RDD[5] and YOLOv7-RDD[7] have reduced reliance on anchor boxes and streamlined architectures, enhancing detection accuracy and efficiency in complex scenarios These modifications have proven effective in front-view and UAV-based road inspections, addressing diverse road conditions. The incorporation of attention mechanisms and feature fusion has enhanced model robustness. Methods such as ensemble learning with attention mechanisms[11] improve detection under varying environmental conditions, while local sensing networks like Lsf-rdd[10] enhance multi-scale feature fusion for detecting various damage types. To achieve real-time detection, models like FPDDN[8] use deformable transformers and lightweight modules, providing fast and accurate detection of irregular damages. LMFE-RDD[6] focuses on efficient feature extraction with minimal computational resources, making it suitable for deployment on resource-constrained devices.

These innovations collectively advance the field of road damage detection, enhancing the ability to accurately and efficiently detect damages in diverse and challenging conditions. However, challenges remain in balancing detection speed and accuracy, especially in complex environments with varying scales of damage. To address these challenges, we propose a novel approach—DAPONet (Dual Attention and Partially Overparameterized Network). DAPONet introduces a dual attention mechanism to integrate global and local information and a multi-scale partial over-parameterization module to handle different scales of damage effectively. Additionally, an efficient downsampling module is incorporated to optimize computational efficiency. 
The main contributions of this study are as follows:

\begin{figure*}[ht]
    \centering
    \includegraphics[width=0.92\linewidth]{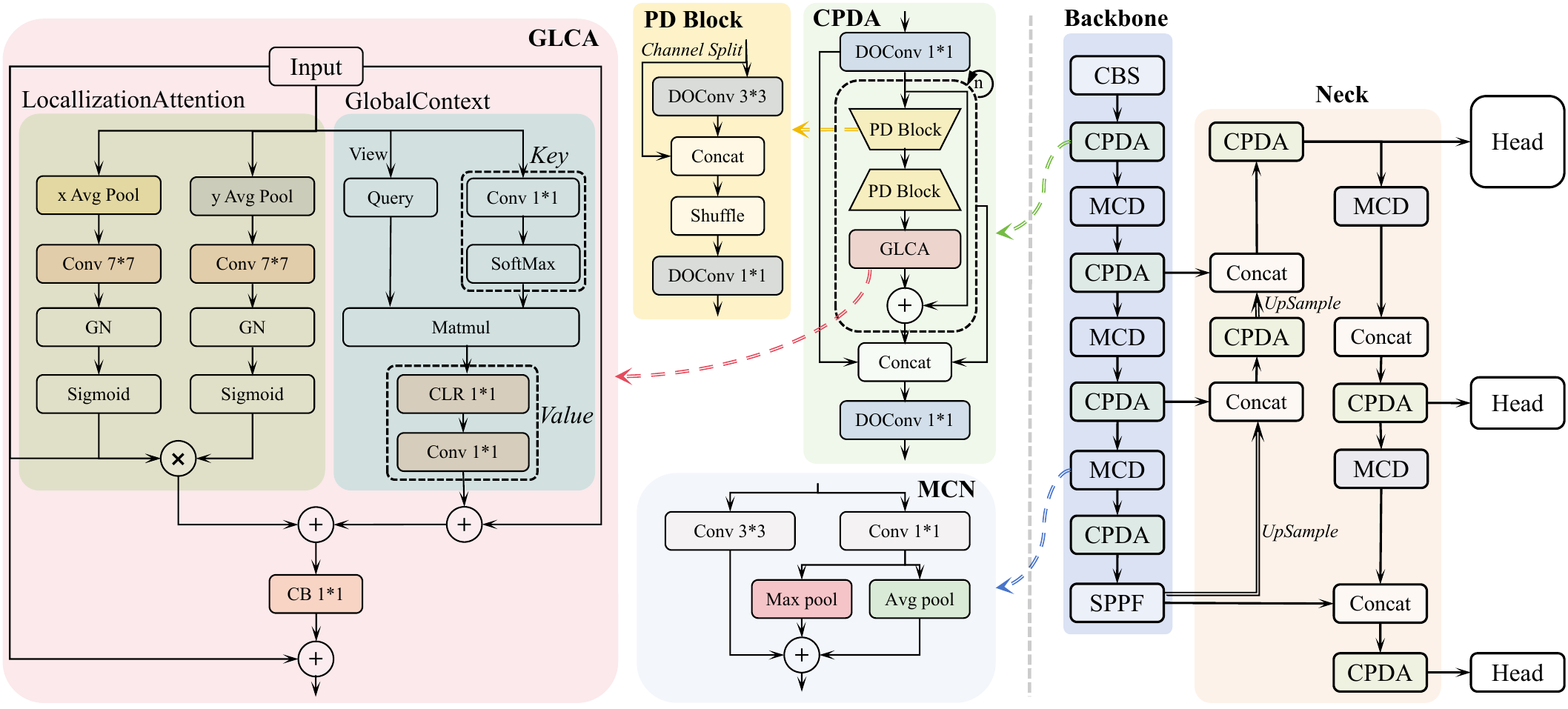}
    \caption{Shows the overall framework. In GLCA  Module, GN is Group Normalization, CLR is Conv-Layer Normal-ReLU, and CB is Conv-Batch Normalization. In the PB Block of CPDA  Module, a quarter of the input feature map is divided into DOConv channels, and the rest is directly concat with the feature map after DOConv operation.}
    \label{fig1}
\end{figure*}

\begin{itemize}
\item \textbf{Global Localization Context Attention (GLCA):} This module enhances the model's ability to detect complex backgrounds and multi-scale targets by combining local and global attention mechanisms.

\item \textbf{Cross Stage Partial Depthwise Over-parameterized Attention (CPDA):} CPDA efficiently processes multi-scale features using partial over-parameterized convolution and contextual attention, improving detection accuracy and computational efficiency.

\item \textbf{Mix Convolutional Downsampling (MCD):} MCD downscales and processes feature maps through multiple parallel paths, enhancing feature extraction versatility and efficiency.

\item \textbf{DAPONet (Dual Attention and Partially Overparameterized Network):} Designed for real-time road damage detection in complex scenes, DAPONet excels in multi-scale feature extraction and fusion through dual attention, partial over-parameterization, and parallel downsampling. Validated on SVRDD and MS COCO datasets, DAPONet demonstrates superior performance.
\end{itemize}

\section{Methods}
This section presents an overview of the proposed model, detailing each module's structure and function. We begin with a general explanation of the model, followed by in-depth descriptions of the key modules: Global Localization Context Attention (GLCA), Cross Stage Partial Depthwise Over-parameterized Attention (CPDA), and Mix Convolutional Downsampling (MCD).



\subsection{Overview}
DAPONet is designed to address the challenges of road damage detection by combining two primary components: the Backbone and the Neck, culminating in the Head for generating detection results. The Backbone is responsible for initial feature extraction, utilizing CPDA (Cross Stage Partial Depthwise Over-parameterized Attention) and MCD (Mix Convolutional Downsampling) modules. These modules enhance the expressiveness of the model by capturing complex patterns and reducing computational overhead through effective downsampling techniques. The Neck further processes these extracted features, employing upsampling, downsampling, and concatenation to ensure that features from different scales are integrated effectively. 

\subsection{Global Localization Context Attention (GLCA) Module}

The GLCA module plays a critical role in DAPONet by combining local and global attention mechanisms to improve detection accuracy in complex environments. It consists of two sub-modules: Efficient Localization Attention (ELA) and Global Context Block (GC). ELA targets local feature extraction by using directional pooling and convolution operations to create attention weights that emphasize key regions within the image. This helps the model focus on significant local details such as small cracks or subtle damages. The GC sub-module, on the other hand, captures broader contextual information by transforming feature maps into a global perspective through matrix multiplication. This process is followed by non-linear activation and normalization, ensuring that the model maintains a balance between detailed local features and overall global context, enhancing its ability to interpret complex scenes.

\subsection{Cross Stage Partial Depthwise Over-parameterized Attention (CPDA) Module}

The CPDA module is designed to enhance the model's ability to process multi-scale features efficiently. It uses DOConv\cite{b7} , which increases feature diversity and improves the representation power of the model. Within the CPDA, PD Blocks are used to capture and expand the receptive field, allowing the model to understand complex contextual information better. The CPDA further integrates the output from the GLCA, ensuring that both local and global contexts are considered. This integration is crucial for accurately detecting road damages of varying sizes and shapes. The final output of the CPDA is a condensed feature map, achieved through concatenation and additional convolution, which not only reduces dimensionality but also integrates diverse feature information efficiently.

\subsection{Mix Convolutional Downsampling (MCD) Module}

The MCD module addresses the need for efficient feature extraction across multiple scales. It does so by processing the input feature maps through multiple parallel paths that utilize different convolution and pooling techniques. Each path focuses on capturing specific details: some paths emphasize edge information and fine details, while others focus on broader, structural features. By combining these paths, the MCD module manages resolution effectively and ensures that critical features are preserved even after downsampling. This multi-path approach reduces computational complexity while maintaining high detection accuracy, making the MCD module a key component in enhancing the versatility and efficiency of DAPONet.

\begin{figure*}[ht]
    \centering
    \includegraphics[width=0.85\linewidth]{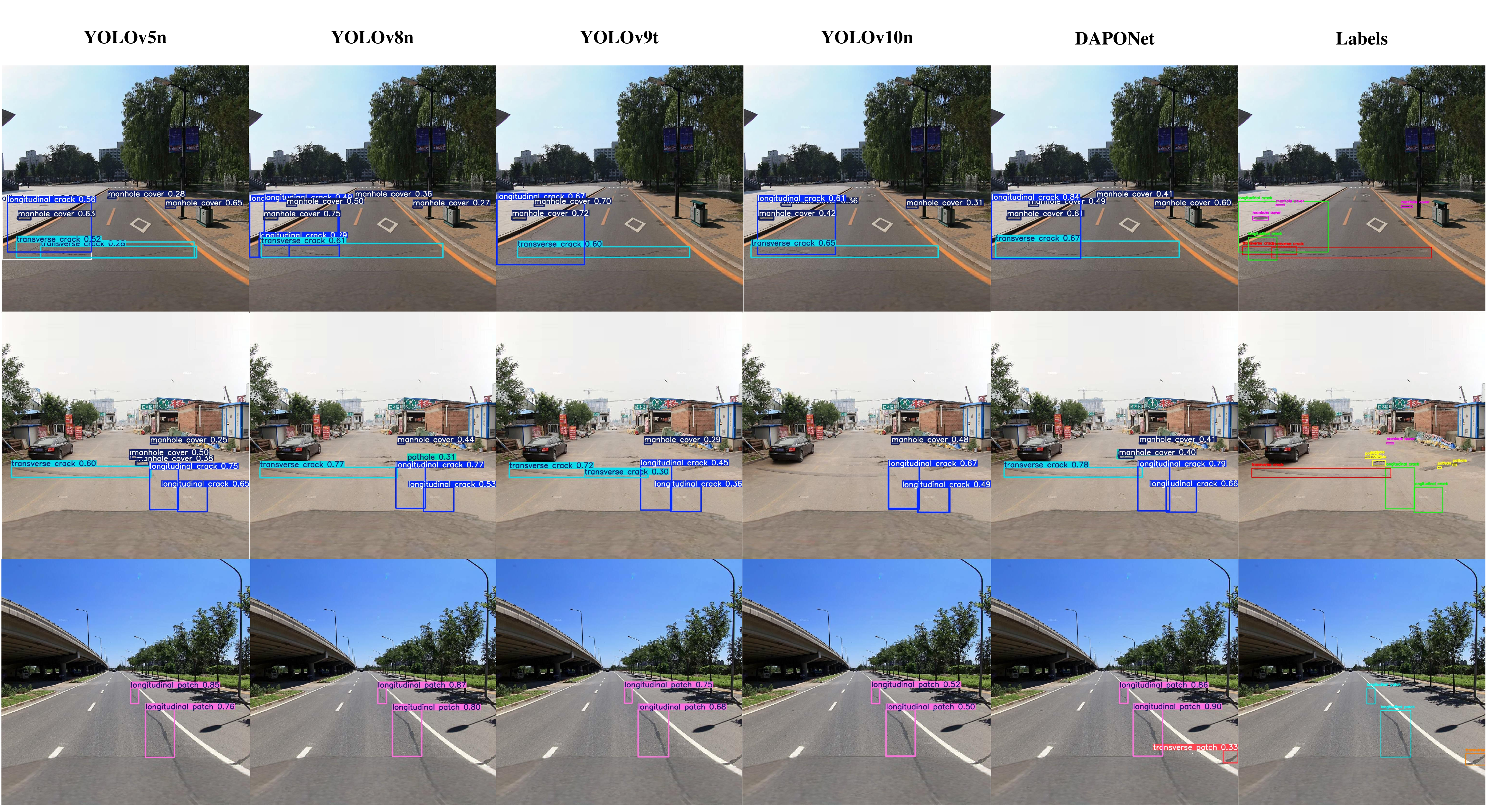}
    \caption{Experimental models recognize visual results on the SVRDD dataset. Different models vary in detecting road damage. YOLOv5n struggles with minor cracks, resulting in higher miss rates. YOLOv8n improves accuracy for transverse cracks and manhole covers but still misses subtle damages. YOLOv9t prioritizes speed but loses precision in detecting finer details. YOLOv10n is effective for larger cracks but has more false negatives for smaller damages. DAPONet outperforms the other models, accurately detecting a broad range of damages, including fine cracks and manhole covers, with high confidence and fewer errors, demonstrating its robustness and precision in various road damage scenarios.
}
    \label{fig2}
\end{figure*}

\section{Experimental details}
In this section, a brief overview of the experimental setup and related resources is presented. Next, the experimental dataset, the experimental setup and the evaluation metrics are presented in turn.

\subsection{Datasets}
1) SVRDD (Street View Image Dataset for Automated Road Damage Detection) dataset: The SVRDD dataset\cite{b8} is a pioneering street view image dataset for road damage detection.

2) The MS COCO (Microsoft Common Objects in Context) dataset: The MS COCO dataset \cite{b9} is a widely recognized benchmark in computer vision.

\subsection{Experimental environment}
The experiments were conducted on a Windows 11 system with an NVIDIA GeForce RTX 3090 GPU, using PyTorch with CUDA 11.8, version 2.0.1, in Jupyter Notebook with Python 3.8. All algorithms were tested under identical conditions. Images were resized to 640 × 640 × 3, with a batch size of 32. The optimizer used was SGD, with a learning rate of 0.001, and the training lasted for 300 epochs.

\subsection{Evaluation metrics}
Four key metrics were used to evaluate model performance: Precision, Recall, mAP50, and MAP50-95\cite{b10} In addition, the lightweight performance of the model is evaluated by FLOPs and Parameters.

\section{Experimental results and discussion and analysis}
In order to validate the superior performance of the DAPONet object detection model proposed in this paper, a series of validations are conducted on the above dataset and several evaluation metrics mentioned above are used for evaluation and analysis.

\subsection{Comparative experiments}

\begin{table}[h]
\centering
\caption{SVRDD dataset test set comparison experiment results}
\begin{tabular}{ c c c c c c }
\toprule
Model  & mAP50  & mAP 50-95  & Params & FLOPs \\
\midrule
YOLOv5n\cite{b11}& 61.7 \% & 35.8 \%& 2.5 M& 7.1 G \\

YOLOv8n\cite{b12}  & 64.5 \% & 37.8 \%& 3.0 M& 8.1 G \\

YOLOv9t\cite{b13}  & 60.8 \% & 36.1 \%& 2.0 M& 7.6 G \\

YOLOv10n\cite{b14}  & 59.7 \% & 35.8 \%& 2.7 M& 8.2 G \\

\textbf{\textbf{DAPONet}} & \textbf{\textbf{70.1 \%}} & \textbf{\textbf{42.8 \%}} & \textbf{\textbf{1.6 M}} & \textbf{\textbf{1.7 G}} \\
\bottomrule
\end{tabular}
\end{table}

In tests on the SVRDD dataset, DAPONet demonstrates superior performance over other real-time object detection models, such as the YOLO series. It excels in precision (71.6\%), recall (66.6\%), and average precision (mAP50: 70.1\%, mAP50-95: 42.8\%), despite having only 1.6M parameters and 1.7G FLOPs, the lowest among all models. This highlights DAPONet’s efficiency and effectiveness in complex, multi-scale damage detection tasks.

\subsection{Generalized object detection experiments}

On the validation set of the MS COCO2017 dataset, DAPONet also demonstrated excellent performance, significantly outperforming the other comparison models. 

\begin{table}[h]
\centering
\caption{Experimental results on MS COCO2017 dataset val}
\begin{tabular}{ c  c  c  c  c  c }
\toprule
Model & mAP 50-95 & Params& FLOPs&  \\
\midrule
NanoDet-Plus-m-1.5x\cite{b15} & 29.9\% & 1.75M & 2.44 G \\

DPNet\cite{b16} & 29.6\% & 1.04 M & 2.5 G   \\

PP-PicoDet-ShuffleNetV2\cite{b17} &30.0\% & 1.17 M & 1.53 G \\

PP-PicoDet-S\cite{b17} & 30.6 \%& \textbf{\textbf{0.99 M}} & \textbf{1.24 G}  \\
EfficientDet-D1\cite{b19} &32.6\% & 6.1 M & 6.6 G   \\
YOLOv5n\cite{b11} & 28.0 \%& 1.9 M & 4.5 G \\
\textbf{\textbf{DAPONet (Ours)}} & \textbf{\textbf{33.4\%}} & 1.6 M & 1.7 G\\
\bottomrule
\end{tabular}
\end{table}

DAPONet achieved 48.3\% of the mAP50 and 33.4\% of the mAP50-95, which is the highest detection accuracy among all the models. These two key metrics far outperformed the other models. Although DAPONet has 1.6M parameters and 1.7G FLOPs, which is slightly higher than some lightweight models, it performs even better in terms of performance. In addition, DAPONet's model size of only 3.6MB is lighter than most models, which further enhances its adaptability in resource-constrained environments such as mobile devices and embedded systems. In contrast, models like NanoDet-Plus-m-1.5x and DPNet, although lower in terms of number of parameters and FLOPs, struggle to match DAPONet in terms of detection accuracy, while EfficientDet-D1, despite being much higher than DAPONet in terms of FLOPs and model size, only has a 32.6\% mAP50-95 which indicates a limited performance improvement despite the significant increase in computational resource consumption.

\subsection{Ablation study}

\begin{table}[h]
\caption{Results of ablation experiments for the SVRDD DATASET TEST SET, where the baseline model is YOLOv8n}
\begin{center}
\begin{tabular}{ c  c  c  c  c  c    }
\toprule
CPDA & MCD &  mAP50 & mAP 50-95  & Params & FLOPs  \\
\midrule
  &    & 64.5 \% & 37.8 \% & 3.0 M & 8.1 G  \\
$\checkmark$ &    & 66.1 \% & 39.2 \% & 2.3 M & 4.6 G  \\
  & $\checkmark$  & 65.2 \% & 38.3 \% & 2.8 M & 7.8 G  \\
$\checkmark$ & $\checkmark$ & \textbf{\textbf{70.1 \%}} & \textbf{\textbf{42.8 \%}} & \textbf{\textbf{1.6 M}} & \textbf{\textbf{1.7 G}}\\
\bottomrule
\end{tabular}
\end{center}
\end{table}
The results of ablation experiments on MS COCO2017 and SVRDD test sets show that the CPDA and MCD modules in DAPONet play a key role in significantly improving the model performance. While the performance of the benchmark model YOLOv8n is good, with the introduction of the CPDA module, the model's recall improves from 59.3\% to 62.1\%, the mAP50 improves from 64.5\% to 66.1\%, and the computational resource requirement is significantly reduced. When the MCD module is introduced, although the precision and recall are also improved, the combination of the two results in the best model performance, with the precision increasing to 71.6\%, the recall reaching 66.6\%, and the mAP50 and mAP50-95 reaching 70.1\% and 42.8\%, respectively. At the same time, the number of parameters and FLOPs of the model are greatly reduced, and the model size is reduced to 3.7MB.

\subsection{Error analysis}
The error analysis for DAPONet shows that while the model excels in detecting a wide range of road damages with high accuracy, it still faces challenges. The primary errors include occasional false positives, where non-damage features like shadows or surface marks are mistakenly identified as damages, and false negatives, particularly with very fine cracks or minor patches under low-contrast conditions. Localization errors, where the model inaccurately predicts the size or location of damage, also occur but are less frequent. These errors suggest the need for further refinement in feature discrimination and sensitivity to subtle details, as well as improvements in handling diverse environmental conditions to enhance DAPONet's overall robustness and reliability in real-world applications.

\section{Conclusion}

In this paper, we present DAPONet, an object detection model designed for SVRDD tasks. By introducing a combination of multi-level and multi-modules, including key modules such as CPDA, MCD, and GLCA, DAPONet exhibits excellent capabilities in feature extraction, feature fusion, and global contextual information integration. Experimental results show that DAPONet significantly outperforms existing mainstream detection models on both SVRDD and MS COCO datasets. Although DAPONet has achieved significant performance improvement in road damage detection, it still suffers from problems such as missed detection as well as wrong detection. In the future, we will consider combining techniques such as Generative Adversarial Networks (GAN) to generate more training data from different scenarios to enhance the model's adaptability to different scenarios as well as unknown environments.


\begin{thebibliography}{00}

\bibitem{b1} L. Fan, D. Cao, C. Zeng, B. Li, Y. Li, and F.-Y. Wang, "Cognitive-based crack detection for road maintenance: An integrated system in cyber-physical-social systems," in IEEE Transactions on Systems, Man, and Cybernetics: Systems, vol. 53, no. 6, pp. 3485-3500, June 2023, doi: 10.1109/TSMC.2022.3227209.

\bibitem{b2} Y. Yuan, M. S. Islam, Y. Yuan, S. Wang, T. Baker, and L. M. Kolbe, "EcRD: Edge-Cloud Computing Framework for Smart Road Damage Detection and Warning," in IEEE Internet of Things Journal, vol. 8, no. 16, pp. 12734-12747, 15 Aug. 2021, doi: 10.1109/JIOT.2020.3024885.

\bibitem{b3} O. Iparraguirre, N. Iturbe-Olleta, A. Brazalez, and D. Borro, "Road marking damage detection based on deep learning for infrastructure evaluation in emerging autonomous driving," in IEEE Transactions on Intelligent Transportation Systems, vol. 2 

\bibitem{b4} D. Ma, H. Fang, N. Wang, C. Zhang, J. Dong, and H. Hu, "Automatic Detection and Counting System for Pavement Cracks Based on PCGAN and YOLO-MF," in IEEE Transactions on Intelligent Transportation Systems, vol. 23, no. 11, pp. 22166-22178, Nov. 2022, doi: 10.1109/TITS.2022.3161960.

\bibitem{b5} J. Li, Z. Qu, S.-Y. Wang, and S.-F. Xia, "YOLOX-RDD: A Method of Anchor-Free Road Damage Detection for Front-View Images," in IEEE Transactions on Intelligent Transportation Systems, doi: 10.1109/TITS.2024.3389945.

\bibitem{b6} He, Q., Li, Z.,  Yang, W. (2024). LMFE-RDD: A road damage detector with a lightweight multi-feature extraction network. Multimedia Systems, 30, 176. https://doi.org/10.1007/s00530-024-01367-z

\bibitem{new1}Z. Ning, H. Wang, S. Li and Z. Xu, "YOLOv7-RDD: A Lightweight Efficient Pavement Distress Detection Model," in IEEE Transactions on Intelligent Transportation Systems, vol. 25, no. 7, pp. 6994-7003, July 2024, doi: 10.1109/TITS.2023.3347034. 

\bibitem{new2}Y. Zhang and C. Liu, "Real-Time Pavement Damage Detection With Damage Shape Adaptation," in IEEE Transactions on Intelligent Transportation Systems, doi: 10.1109/TITS.2024.3416508.

\bibitem{new3}J. Zhu, Y. Wu and T. Ma, "Multi-Object Detection for Daily Road Maintenance Inspection With UAV Based on Improved YOLOv8," in IEEE Transactions on Intelligent Transportation Systems, doi: 10.1109/TITS.2024.3437770. 

\bibitem{new4}He, Q., Li, Z.  Yang, W. Lsf-rdd: a local sensing feature network for road damage detection. Pattern Anal Applic 27, 99 (2024). https://doi.org/10.1007/s10044-024-01314-8

\bibitem{new5}S. Wang, H. Jiao, X. Su and Q. Yuan, "An Ensemble Learning Approach With Attention Mechanism for Detecting Pavement Distress and Disaster-Induced Road Damage," in IEEE Transactions on Intelligent Transportation Systems, doi: 10.1109/TITS.2024.3391751. 

\bibitem{new6}Chen, DR., Chiu, WM. Deep-learning-based road crack detection frameworks for dashcam-captured images under different illumination conditions. Soft Comput 27, 14337–14360 (2023). https://doi.org/10.1007/s00500-023-08738-0

\bibitem{b7} J. Cao et al., "DO-Conv: Depthwise Over-Parameterized Convolutional Layer," in IEEE Transactions on Image Processing, vol. 31, pp. 3726-3736, 2022, doi: 10.1109/TIP.2022.3175432.

\bibitem{b8} Ren, M., Zhang, X., Zhi, X. et al. (2024). An annotated street view image dataset for automated road damage detection. Scientific Data, 11, 407. https://doi.org/10.1038/s41597-024-03263-7

\bibitem{b9}Lin, Tsung-Yi, Michael Maire, Serge Belongie, Lubomir Bourdev, Ross Girshick, James Hays, Pietro Perona, Deva Ramanan, C. Lawrence Zitnick, and Piotr Dollar. "Microsoft COCO: Common Objects in Context." arXiv preprint arXiv:1405.0312 (2015). [Online]. Available: https://arxiv.org/abs/1405.0312

\bibitem{b10}Chen, W., Luo, J., Zhang, F. et al. (2024). A review of object detection: Datasets, performance evaluation, architecture, applications and current trends. Multimedia Tools and Applications, 83, 65603–65661. https://doi.org/10.1007/s11042-023-17949-4

\bibitem{b11}Jocher, G. (2020). YOLOv5 by Ultralytics (Version 7.0) [Computer software]. https://doi.org/10.5281/zenodo.3908559

\bibitem{b12}Glenn Jocher, Ayush Chaurasia, and Jing Qiu. (2023). Ultralytics YOLOv8, Version 8.0.0. [Software]. Available from https://github.com/ultralytics/ultralytics. License: AGPL-3.0.

\bibitem{b13}Wang, Chien-Yao, I-Hau Yeh, and Hong-Yuan Mark Liao. "YOLOv9: Learning What You Want to Learn Using Programmable Gradient Information." arXiv preprint arXiv:2402.13616 (2024). [Online]. Available: https://arxiv.org/abs/2402.13616

\bibitem{b14}Wang, Ao, Hui Chen, Lihao Liu, Kai Chen, Zijia Lin, Jungong Han, and Guiguang Ding. "YOLOv10: Real-Time End-to-End Object Detection." arXiv preprint arXiv:2405.14458 (2024). [Online]. Available: https://arxiv.org/abs/2405.14458

\bibitem{b15}RangiLyu. (2021). NanoDet-Plus: Super fast and high accuracy lightweight anchor-free object detection model. [Software]. Available from https://github.com/RangiLyu/nanodet.

\bibitem{b16}Q. Zhou, H. Shi, W. Xiang, B. Kang, and L. J. Latecki, "DPNet: Dual-Path Network for Real-Time Object Detection With Lightweight Attention," in IEEE Transactions on Neural Networks and Learning Systems, doi: 10.1109/TNNLS.2024.3376563.

\bibitem{b17}Yu, Guanghua, Qinyao Chang, Wenyu Lv, Chang Xu, Cheng Cui, Wei Ji, Qingqing Dang, Kaipeng Deng, Guanzhong Wang, Yuning Du, Baohua Lai, Qiwen Liu, Xiaoguang Hu, Dianhai Yu, and Yanjun Ma. "PP-PicoDet: A Better Real-Time Object Detector on Mobile Devices." arXiv preprint arXiv:2111.00902 (2021). [Online]. Available: https://arxiv.org/abs/2111.00902


\bibitem{b19}M. Tan, R. Pang, and Q. V. Le, "EfficientDet: Scalable and Efficient Object Detection," in 2020 IEEE/CVF Conference on Computer Vision and Pattern Recognition (CVPR), Seattle, WA, USA, 2020, pp. 10778-10787, doi: 10.1109/CVPR42600.2020.01079.



\end{thebibliography}
\end{document}